\documentclass[a4paper,10pt]{cas-dc}

\usepackage{threeparttable}
\usepackage[round]{natbib}
\usepackage{amsfonts,amssymb}
\usepackage{amsmath}
\usepackage{stfloats}
\usepackage{setspace}
\usepackage{subfigure}
\usepackage{cleveref}
\usepackage{graphicx}
\usepackage{caption}
\usepackage{balance}
\usepackage{graphicx} 
\usepackage{caption}
\usepackage{array} 
\usepackage{booktabs} 
\usepackage{multirow}
\usepackage{bbding}
\usepackage{booktabs}
\usepackage{hyperref}
\usepackage{bm}

\graphicspath{./PDF/}
\DeclareGraphicsExtensions{.pdf, .jpeg, .png, .jpg, .eps}
\definecolor{citeref}{RGB}{10,110,180} 
\hypersetup{hypertex=true,
	colorlinks=true,
	linkcolor=citeref,
	anchorcolor=citeref,
	urlcolor=citeref,
	citecolor=citeref}

\begin{document}
	
\let\WriteBookmarks\relax
\def\floatpagepagefraction{1}
\def\textpagefraction{.001}
\let\printorcid\relax

\shorttitle{MCFNet: A Multimodal Collaborative Fusion Network for Fine-Grained Semantic Classification}


\shortauthors{Y. Qiao et al.}  
\title[mode = title]{MCFNet: A Multimodal Collaborative Fusion Network for Fine-Grained Semantic Classification}  

\author{Yang Qiao}
\ead{6231922005@stu.jiangnan.edu.cn}
\credit{Methodology, Data curation, Formal analysis, Writing – original draft}

\author{Xiaoyu Zhong}
\ead{7221923002@stu.jiangnan.edu.cn}
\credit{Conceptualization, Software}

\author{Xiaofeng Gu}
\ead{xgu@jiangnan.edu.cn}
\credit{Conceptualization, Supervision, Funding acquisition, Writing – review and editing}

\author{Zhiguo Yu}
\cormark[1]
\ead{yuzhiguo@jiangnan.edu.cn}
\credit{Conceptualization, Project administration, Supervision, Writing – review and editing}

\address{School of Integrated Circuits, Jiangnan University, Wuxi 214122, China}

\cortext[1]{Corresponding author} 

\begin{abstract}
Multimodal information processing has become increasingly important for enhancing image classification performance. However, the intricate and implicit dependencies across different modalities often hinder conventional methods from effectively capturing fine-grained semantic interactions, thereby limiting their applicability in high-precision classification tasks.
To address this issue, we propose a novel Multimodal Collaborative Fusion Network (MCFNet) designed for fine-grained classification. The proposed MCFNet architecture incorporates a regularized integrated fusion module that improves intra-modal feature representation through modality-specific regularization strategies, while facilitating precise semantic alignment via a hybrid attention mechanism. Additionally, we introduce a multimodal decision classification module, which jointly exploits inter-modal correlations and unimodal discriminative features by integrating multiple loss functions within a weighted voting paradigm. Extensive experiments and ablation studies on benchmark datasets demonstrate that the proposed MCFNet framework achieves consistent improvements in classification accuracy, confirming its effectiveness in modeling subtle cross-modal semantics.
\end{abstract}




\maketitle

\section{Introduction}
\label{Section 1}

Fine-grained image classification aims to accurately distinguish subcategories within a broader category and has been widely applied in domains, such as biodiversity monitoring~\citep{tuia2022perspectives}, disease diagnosis~\citep{fu2024cafnet}, cellular analysis~\citep{rubin2019top}, and weapon identification~\citep{chen2024deep}. However, traditional single-modal classification approaches often struggle to capture subtle inter-class variations due to limited semantic information and insufficient data, resulting in degraded performance.

\begin{figure}[t]
	\centering
	 \subfigure[Traditional Single-Modal Image Classification]{
        \includegraphics[scale=0.53]{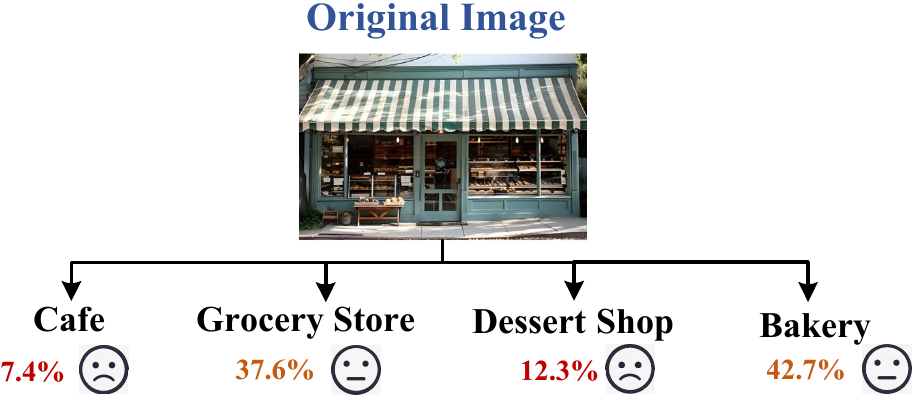}
        \label{Figure 1-(a)}
    }
    \vspace{0.5cm} 
    \subfigure[Multimodal Fine-Grained Image Classification]{
        \includegraphics[scale=0.65]{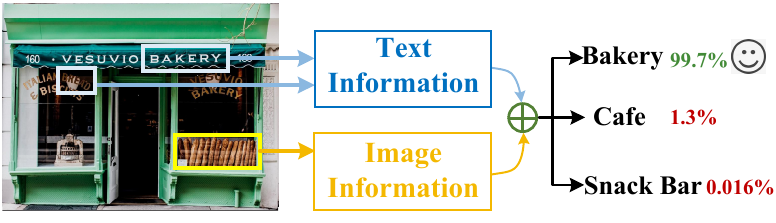}
        \label{Figure 1-(b)}
    }
     \vspace{-0.5cm}
    \caption{Comparison of image classification results. (a) Image classification using only image unimodal data results in lower classification accuracy. (b) Utilizing multimodal data for image classification significantly enhances classification accuracy.}
    \label{Figure 1}
    \vspace{-0.5cm}
\end{figure}

As shown in Fig.~\ref{Figure 1}, unimodal image classification methods (Fig.~\ref{Figure 1-(a)}) tend to overlook fine-grained cues, such as slight differences in shape, texture, or contextual semantics, which are critical for distinguishing visually similar categories. In contrast, integrating complementary modalities, such as textual descriptions or metadata, provides richer semantic context and significantly enhances classification performance (Fig.~\ref{Figure 1-(b)}).
To mitigate the limitations of unimodal learning, recent research has focused on multimodal fusion techniques, which combine visual and linguistic representations for more robust fine-grained understanding~\citep{prakash2021multi,xu2021multimodal,fu2022cma}. Existing methods can be broadly categorized into two groups: traditional hand-crafted fusion techniques and deep learning-based approaches. Hand-crafted strategies, such as canonical correlation analysis (CCA)~\citep{low2004distinctive} and partial least squares (PLS)~\citep{he2008scale}, are computationally efficient and easy to implement. However, they often rely on extensive annotated features with strong discriminative power~\citep{xu2023improving}, resulting in high manual labeling costs.

Deep learning-based methods overcome these limitations by enabling automatic feature extraction and fusion~\citep{li2024research,ji2023dual,huang2024fusiondiff}. Nevertheless, challenges remain. Graph neural networks (GNNs)~\citep{zhou2020graph}, for instance, require domain-specific prior knowledge to construct graph structures, which limits their generalization capability. Similarly, densely connected convolutional networks~\citep{huang2017densely} lack architectural flexibility in multimodal settings and demand significant expertise for effective training.

Recently, multimodal pre-trained models have shown strong generalization ability by aligning vision-language representations through contrastive learning~\citep{li2023fine,yang2025omnidialog}. These models offer a scalable solution to enhance fine-grained semantic understanding across modalities.
Motivated by these advancements, we propose a novel {Multimodal Collaborative Fusion Network (MCFNet)} for fine-grained image classification. Our framework incorporates intra-modal regularization strategies to strengthen single-modal feature representations, a regularized integrated fusion module for semantic alignment, and a multimodal decision classification module that leverages weighted voting to improve prediction robustness.
The main contributions of this paper are summarized as follows:
\begin{itemize}
    \item[$\bullet$] We propose MCFNet, a collaborative fusion framework for fine-grained image classification that facilitates deep cross-modal interactions by integrating multi-dimensional visual and textual features.
    \item[$\bullet$] We introduce a regularized integrated fusion module that combines diverse regularization techniques with a hybrid attention mechanism, enhancing semantic alignment across modalities.
    \item[$\bullet$] We design a multimodal decision classification module that adopts multiple loss functions and a weighted voting strategy to improve classification accuracy and model generalization.
    \item[$\bullet$] Extensive experiments on the Con-Text and Drink Bottle datasets demonstrate the effectiveness of MCFNet. Ablation studies validate the contributions of our hybrid attention, regularization modules, and composite loss functions.
\end{itemize}

The remainder of this paper is organized as follows. Section~\ref{Section 2} reviews related work. Section~\ref{Section 3} details the proposed MCFNet framework. Experimental settings and results are provided in Section~\ref{Section 4}. Section~\ref{Section 5} concludes the paper and outlines future directions.

\section{Related Work}
\label{Section 2}
Recent advances in fine-grained image classification have introduced a range of deep learning-based strategies. In this section, we categorize relevant literature into three major directions: (1) region-based and end-to-end feature encoding methods, (2) transformer-based architectures, and (3) multimodal and externally enhanced approaches.

\subsection{Feature Encoding Approaches}
Early approaches to fine-grained classification emphasized identifying subtle, discriminative regions for accurate recognition. Recurrent attention networks~\citep{fu2017look}, multi-attention frameworks~\citep{sun2018multi}, and trilinear attention mechanisms~\citep{zheng2019looking} have been proposed to localize critical patches. However, these methods often require two-stage pipelines and suffer from weak interaction between localization and classification modules.
To simplify this, end-to-end architectures such as bilinear pooling networks~\citep{lin2015bilinear}, compact and low-rank variants~\citep{gao2016compact,kong2017low}, and kernel-based pooling~\citep{cui2017kernel} have emerged, learning global representations directly from raw inputs. Other efforts focus on enhancing feature discriminability, such as interactive channel loss~\citep{chang2020devil} and counterfactual attention learning~\citep{rao2021counterfactual}. More recently, Shen et al.~\citep{shen2025imaggarment} proposed IMAGGarment-1, which applies controllable feature encoding for fine-grained garment generation, revealing that structure-aware conditioning can further enhance recognition in fashion-related scenarios.

\subsection{Transformer-Based Methods}
The introduction of Vision Transformers (ViT)~\citep{dosovitskiy2020image} enabled the capture of long-range dependencies, offering an effective alternative to convolution-based models. Extensions like Swin Transformer~\citep{liu2021swin} and TransFG~\citep{he2022transfg} improve local-global feature integration through hierarchical attention and part-aware modeling. Advanced ViT variants, such as RAMS~\citep{hu2021rams} and MixFormer~\citep{yu2023mix}, further enhance scale-adaptivity and generalization under limited data.
To bridge the gap between structure and attention, Shen and Tang~\citep{shen2024imagpose} proposed IMAGPose, a unified conditional framework that leverages pose guidance within a Transformer backbone for more precise person generation. Their approach highlights the effectiveness of condition-aware attention in learning subtle appearance variations, which is crucial for fine-grained tasks.

\subsection{Multimodal and External Information}
Due to limited labeled samples in fine-grained datasets, recent works have explored leveraging external information or multimodal inputs. Web data~\citep{huang2021toan}, textual descriptions~\citep{wang2022knowledge}, and vision-language alignment frameworks~\citep{jiang2023cross} have been employed to enrich semantic context. Fine-tuning large vision-language models with instruction datasets, such as in Pink~\citep{xuan2024pink}, further improves multimodal understanding. He et al.~\citep{he2025analyzing} additionally introduce pose cues to enhance vision-text alignment.
Building on this trend, Shen et al.~\citep{shen2025imagdressing} proposed IMAGDressing-v1 for customizable virtual dressing, combining visual, textual, and pose conditions. In parallel, they extended diffusion models to model motion-aware generation~\citep{shen2025long} and story visualization~\citep{shen2025boosting}, demonstrating the effectiveness of rich contextual modeling in multimodal generation tasks~\citep{shen2023advancing}. These studies underscore the potential of contrastive pretraining and conditional fusion for cross-modal representation learning.

While previous works have demonstrated progress in region discovery, global feature learning, and multimodal alignment, challenges remain in designing unified, generalizable, and efficient frameworks. In this work, we propose a collaborative multimodal network that leverages pre-trained encoders, hybrid attention, and lightweight regularization to address these limitations in fine-grained classification.

\begin{figure*}[t]
	\centering
	\includegraphics[scale=0.85
	]{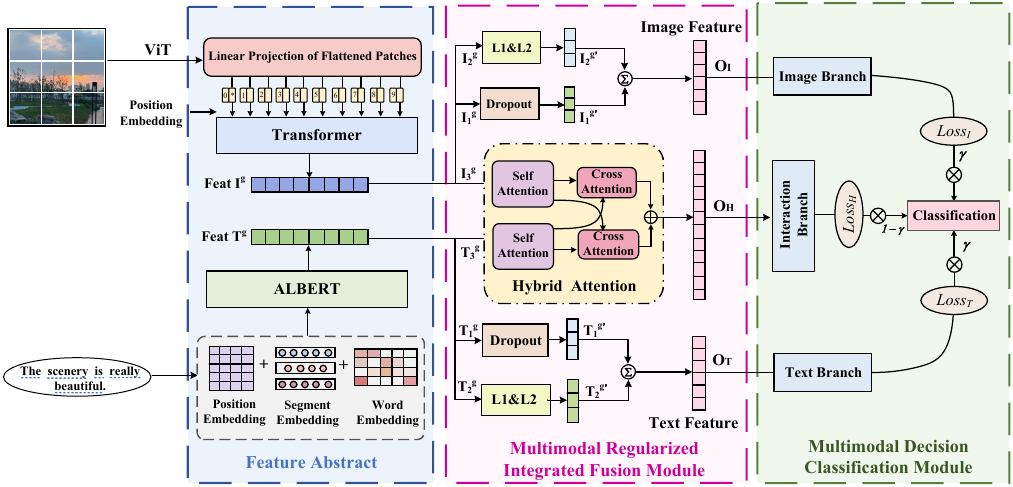}
	\caption{Detailed network architecture. The network employs pre-trained models ALBERT and ViT to extract textual and visual information, respectively. Subsequently, a multimodal regularization integration fusion network is used to fuse the features of scene text regions with those of visually salient objects. In the multimodal decision classification network, the network utilizes multiple loss functions and incorporates a dynamic weight adjustment mechanism to achieve precise fine-grained classification.}
	\label{Figure 2}
\end{figure*}

\section{Methods}
In this section, we propose a fine-grained classification framework (MCFNet) and introduce its overall architecture. The network comprises three key components: a feature extractor, a regularized integrated fusion module, and a decision classification module. We elaborate on the network structure and the mechanisms of these three components in detail. Finally, we present the comprehensive loss function employed by MCFNet. 

\label{Section 3}
\subsection{Overall Framework}
The overall framework of MCFNet is shown in Figure \ref{Figure 2}. This network architecture comprises three components: feature extraction, multimodal regularized feature fusion, and multimodal decision classification. In the feature extraction stage, the text modality utilizes the ALBERT model, which employs a multi-layer transformer to process and obtain text feature representations $T_1^g$, $T_2^g$, and $T_3^g$. The image modality employs the ViT model to serialize image patches, which are then encoded through the transformer to produce $I_1^g$, $I_2^g$, and $I_3^g$. In the multimodal regularization feature fusion stage, we apply Dropout and regularization techniques to the text features $T_1^g$, $T_2^g$ and the image features $I_1^g$, $I_2^g$, in order to enhance the expressive capability of individual modal features. Meanwhile, a hybrid attention mechanism is applied to the text feature $T_3^g$ and the image feature $I_3^g$ for feature fusion, resulting in the multimodal fusion feature $O_H$. In the multimodal decision classification stage, $O_I$, $O_H$, and $O_T$ are processed through independent classification branches and the Softmax activation function to derive the classification results. Additionally, during the network’s training process, multiple loss functions are introduced, with the primary loss being $Loss_H$ corresponding to $O_H$ and the auxiliary losses being $Loss_I$ and $Loss_T$ corresponding to $O_I$ and $O_T$. This approach dynamically adjusts parameters through backpropagation to balance multimodal and unimodal feature learning, thereby enhancing the network's generalization and classification performance.

\subsection{Multimodal feature extractor}
In this paper, we utilize distinct pre-trained models to extract features from different modalities according to their specific characteristics.
\subsubsection{Text Feature Extraction} 
We employ the pre-trained ALBERT \citep{lan2019albert} to extract semantic and contextual information from the text. For each input text $T$, we first utilize the word segmenter to segment the original text, converting it into word-level embeddings $W=\left\{{w_i} \right\}\left| {_{i = 1}^D}\right.$ that can be processed by the word embedding layer. Each word's embedding vector is composed of the sum of its word embedding vector, position embedding vector, and segment embedding vector. Subsequently, we transform $w_i$ into a $h$-dimensional vector by employing an embedding layer. The set of word embeddings is represented as $T^l=\left\{ {t_i^l} \right\}\left| {_{i = 1}^D} \right.$. These sequences are fed into the Transformer to capture contextualized word embeddings. We denote the output as textual context embeddings $T^c=\left\{ {t_i^c} \right\}\left| {_{i = 1}^D} \right.$, where $D$ denotes the number of the words in the sentence. The final global embedding $T^g$ is obtained by aggregating the context embeddings $T^c$.

\subsubsection{Image Feature Extraction}  
We employ a pre-trained ViT \citep{dosovitskiy2020image} to extract features from images. For the given original image $I \in {R^{H \times W \times C}}$, we first split $I$ into a sequence of $N = H \times W/{P^2}$ fixed-sized non-overlapping patches, where $P$ denotes the patch size, and $H$ and $W$ are the height and width. Then we map the patch sequence to 1D tokens $I^i=\left\{ {I_i^v} \right\}\left| {_{i = 1}^N} \right.$ by a trainable linear projection. With injection of positional embedding and extra [CLS] token, the sequence of tokens $\left\{ {I_{cls}^v,I_1^v, \cdots ,I_N^v} \right\}$ are input into L-layer transformer blocks to model correlations of each patch and obtain contextualized object embeddings. The output of the context encoder is denoted as visual context embeddings $I^c=\left\{ {I_i^c} \right\}\left| {_{i = 1}^N} \right.$. Finally, the visual global embedding $I^g$ is obtained by integrating the context embeddings $I^c$.

\subsection{Multimodal Regularized Integrated Fusion Module}
\label{section 3.3} 
The module integrates three distinct regularization constraint methods to address overfitting and enhance the network's generalization capability. Furthermore, it incorporates a hybrid attention mechanism to achieve improved multimodal semantic alignment. 

\subsubsection{Regularization Feature Representations} 
To enhance the network's capacity to distinguish features within text and image modalities, we design different regularization methods for each modality: regularization based on dropout; regularization based on elastic network; and regularization based on hybrid attention network.

Dropout regularization is applied to the image feature $I_1^g$ and the text feature $T_1^g$ to randomly drop neurons in the network to prevent overfitting of the network. The final outputs ${I_1^g}^{'}$ and ${T_1^g}^{'}$ are calculated as follows:
\begin{equation}
{I_1^g}^{'} = \text{Dropout}(I_1^g, p),
\label{eq:L1}
\end{equation}
\begin{equation}
{T_1^g}^{'} = \text{Dropout}(T_1^g, p),
\label{eq:L2}
\end{equation}

\noindent where the \(\text{Dropout}\) function is defined as:
\begin{equation}
\text{Dropout}(x, p) =
\begin{cases}
\displaystyle \frac{x \odot \text{Bernoulli}(p)}{p}, & \text{if training}, \\
x, & \text{if inference},
\end{cases}
\label{eq:L3}
\end{equation}

\noindent where \(x\) denotes the input feature, \(p\) is the dropout probability indicating the likelihood of a neuron being retained, \(\odot\) represents the element-wise multiplication operation, and \(\text{Bernoulli}(p)\) is a Bernoulli distribution used to stochastically generate a binary mask for dropping neurons.

For the image feature $I_2^g$ and the text feature $T_2^g$, elastic network regularization techniques is employed. Elastic network regularization is a method that combines L1 and L2 regularization, achieving a balance between network stability and sparsity by adjusting the weights of the L1 and L2 regularization coefficients. The final outputs ${I_2^g}^{'}$ and ${T_2^g}^{'}$ are calculated as follows:
\begin{equation}
{I_2^g}^{'} = \text{ElasticNet}(I_2^g, \alpha, \beta),
\label{eq:L4}
\end{equation}
\begin{equation}
{T_2^g}^{'} = \text{ElasticNet}(T_2^g, \alpha, \beta),
\label{eq:L5}
\end{equation}

\noindent where the \(\text{ElasticNet}\) function is defined as:
\begin{equation}
\begin{split}
\text{ElasticNet}(x, \alpha, \beta) = \arg\min_{x'} \Big\{ \text{Loss}(x, x') \\
+ \alpha \|x'\|_1 + \beta \|x'\|_2^2 \Big\},
\end{split}
\label{eq:L6}
\end{equation}

\noindent where \(x\) denotes the input feature, \(\alpha\) and \(\beta\) are the regularization coefficients for L1 and L2 regularization respectively. \(\text{Loss} (x, x')\) is the loss function measuring the reconstruction error between the original feature \(x\) and the regularized feature \(x'\). 

The unimodal features derived from the regularization methods based on dropout, elastic network are concatenated. These concatenated features are subsequently passed through a linear layer with an activation function to yield the single-modal fusion features of the two regularization channels.The final outputs $O_{I}$ and $O_{T}$ are calculated as follows:
\begin{equation}
O_{I} = W_1 \cdot \text{Concat}({I_1^g}^{'}, {I_2^g}^{'}) + b_1,
\label{eq:L7}
\end{equation}
\begin{equation}
O_{T} = W_2 \cdot \text{Concat}({T_1^g}^{'}, {T_2^g}^{'}) + b_2,
\label{eq:L8}
\end{equation}

\noindent where \({I_1^g}^{'}\) and \({I_2^g}^{'}\) denote the image features regularized by dropout and elastic net respectively; \({T_1^g}^{'}\) and \({T_2^g}^{'}\) denote the text features regularized by dropout and elastic net respectively. \(\text{Concat}\) represents the feature concatenation operation. \(W_1\) and \(W_2\) are weight matrices, while \(b_1\) and \(b_2\) are bias terms.

\subsubsection{Hybrid Attention Network}  
\begin{figure}[t]
    \raggedright
    \centering
    \subfigure[Interaction Encoder]{
        \includegraphics[scale=0.2]{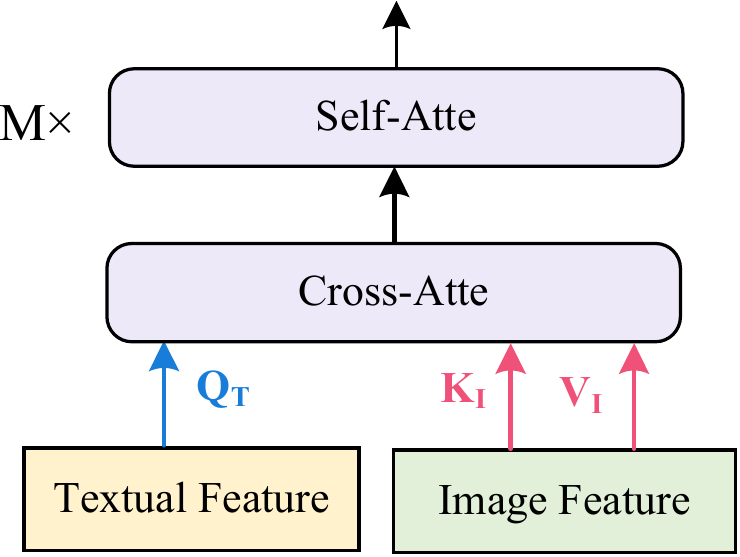}
        \label{Figure 3-(a)}
    }
    \hspace{-0.4cm}  
    \subfigure[Merged Attention]{
        \includegraphics[scale=0.2]{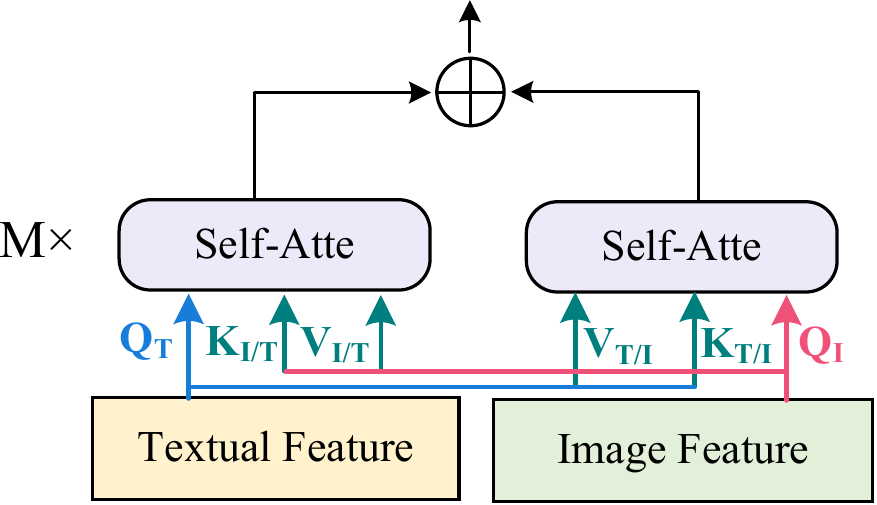}
        \label{Figure 3-(b)}
    }
    \hspace{-0.4cm}  
    \subfigure[Ours]{
        \includegraphics[scale=0.2]{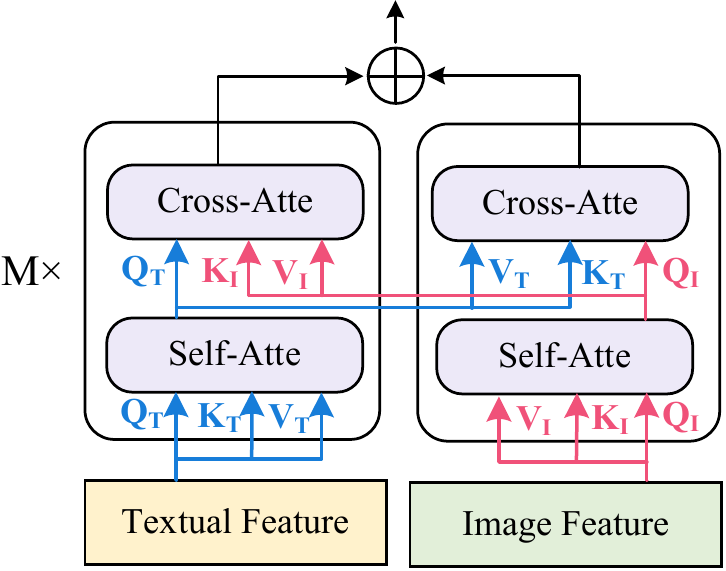}
        \label{Figure 3-(c)}
    }
    \caption{A comparison diagram of our multimodal hybrid attention network with two other popular interaction modules. (a) Interaction encoder. Text and image features are first fused by a cross-attention module, and then fed into a self-attention module. (b) Merged attention. Text and image features are first concatenated and then fed into their respective independent self-attention modules. (c) Our hybrid attention network. Text and image features are first fed into separate self-attention modules, and then sent to a cross-attention module for cross-modal interaction.}
    \label{Figure 3}
\end{figure}

To enhance inter-modal interaction, improve the network's generalization ability, and achieve more precise fine-grained semantic alignment, we design a hybrid attention module, compared to two other popular multimodal attention modules \citep{hendricks2021decoupling,jiang2023cross}, our design is more computationally efficient, as shown in Figure \ref{Figure 3}. This mechanism simultaneously considers self-attention and cross-attention, thereby better capturing the relationships between features when processing complex inputs.

We concatenate the text feature $T_3^g$, extracted by ALBERT, with the image feature $I_3^g$, extracted by ViT. When processing these two unimodal feature vectors, we input them into their respective self-attention modules. This operation aims to ensure that each image region or text word can fully attend to other fine-grained feature elements within the same modality during the attention computation process, thereby uncovering potential associative information within the modality. Specifically, for the image feature $I_3^{\rm{g}} = \left\{ {{i_1},{i_2}, \cdots ,{i_k}} \right\}$, we first utilize the Transformer model to construct the intra-modal relationships among the elements within the image region, thereby obtaining the updated set of image region features $I_{3N}^{\rm{g}} = \left\{ {{i_{N1}},{i_{N2}}, \cdots ,{i_{Nk}}} \right\}$. Subsequently, we aggregate these updated features through an average pooling operation. Finally, we apply a normalization method to the aggregated features to obtain the normalized global representation of the image $I_0$:
\begin{equation}
I_0 = \frac{1}{k} \sum_{i=1}^k i_{Ni}.
\label{eq:L9}
\end{equation}

For the textual feature $T_3^{\rm{g}} = \left\{ {{t_1},{t_2}, \cdots ,{t_k}} \right\}$, we utilize a one-dimensional convolutional neural network to model the local contextual relationships of textual words. Specifically, we construct three different one-dimensional convolutional networks with varying window lengths of 1, 2, and 3, corresponding to the lengths of single words, two-word phrases, and three-word phrases, respectively, in order to explore the contextual associations within phrases of different lengths. The operational process of a one-dimensional convolutional neural network with a window length of 1 for the k-th word can be represented by the following equation:
\begin{equation}
p_{l,n} = \text{ReLU} \left( W_l \cdot t_{n:n+l-1} + b_l \right), \quad l = 1, 2, 3,
\label{eq:L10}
\end{equation}

\noindent where $W_l$ is the weight matrix and $b_l$ is the bias term. 

Subsequently, a max pooling operation is applied to the word features that have completed relational modeling within all modalities, yielding $q_l = \max \left\{ p_{l,1}, \ldots, p_{l,n} \right\}$. The results $q_1$, $q_2$, and $q_3$ are then concatenated and passed into a fully connected layer, followed by normalization to obtain the global embedding feature of the text sentence $T_0$:
\begin{equation}
T_0 = \text{LayerNorm} \left( W_e \cdot\, \text{concat} \left( q_1, q_2, q_3 \right) + b_e \right),
\label{eq:L11}
\end{equation}

\noindent where $W_e \in {R}^{d \times 3d}$ and $b_e \in {R}^{d \times 1}$.

Although the self-attention module has achieved modeling of intra-modal relationships between image regions and text words, it still lacks the capability to model inter-modal relationships between image regions and text words. Therefore, we design a multimodal hybrid network following the self-attention module, which can collaboratively achieve both intra-modal and inter-modal relationship modeling of fine-grained multimodal features within a unified model framework. Specifically, we use the hidden states of the current modality features multiplied by the weight parameter matrix as queries, while the hidden states of the other modality, multiplied by the corresponding weight parameter matrix, serve as keys and values. Subsequently, the fused image features $I_0$ and text features $T_0$ are concatenated and summed. The detailed calculation process is illustrated as follows:
\begin{equation}
I_0^{'}= \text{Softmax}\left( \frac{Q(T_0) K(I_0)^T}{\sqrt{d_k}} \right) V_{I_0},
\label{eq:L12}
\end{equation}
\begin{equation}
T_0^{'} = \text{Softmax}\left( \frac{Q(I_0) K(T_0)^T}{\sqrt{d_k}} \right) V_{T_0},
\label{eq:L13}
\end{equation}
\begin{equation}
O_H = I_0^{'} + T_0^{'},
\label{eq:L14}
\end{equation}

\noindent where $Q$, $K$, and $V$ represent Query, Key, and Value, respectively. The similarity matrix $Q{K^T}$ is generated by performing matrix multiplication between $Q$ and $K^T$. Each element of the similarity matrix is then divided by d, where d is the dimensionality of $K$, which reduces the variance of $Q{K^T}$ and stabilizes gradient updates during training. The Softmax operation normalizes the weight matrix $\frac{{Q{K^T}}}{{\sqrt {{d_k}} }}$ to compute the attention weights. Subsequently, the normalized weight matrix is multiplied by the matrix $V$, and through a weighted summation operation, the enhanced representation $O_H$ is ultimately obtained.

\begin{figure}[t]
	\centering
	\includegraphics[scale=0.55
	]{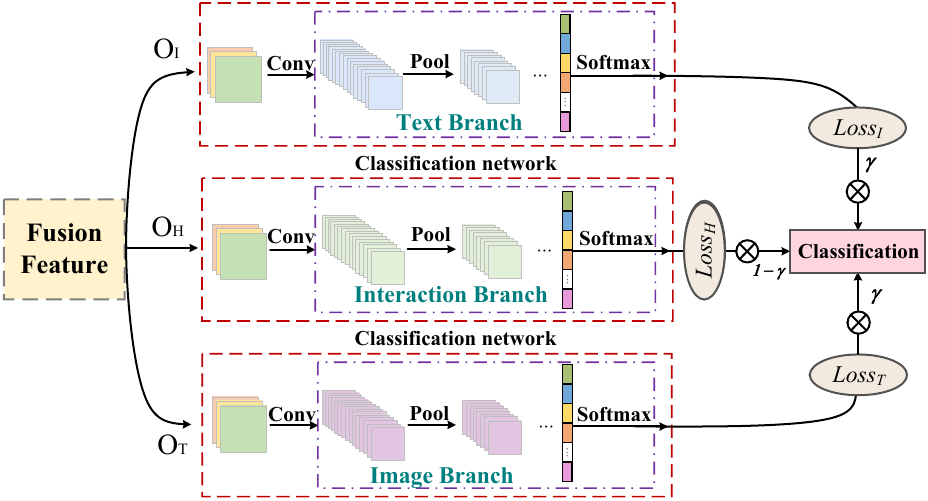}
	\caption{Overview of the multimodal decision classification network. Integrated features $O_T$, $O_H$, and $O_I$ are fed into corresponding text, interaction, and image branches. Each branch contains convolution, pooling, and fully connected layers followed by Softmax. Final predictions are fused via weighted voting with learnable coefficients, and the overall loss is computed accordingly. }
	\label{Figure 4}
    \vspace{-0.5cm}
\end{figure}

\subsection{Multimodal Decision Classification Module}
\label{section 3.4} 
In this study, we construct a dual-driven optimization framework. Decision-level fusion is achieved through a modality-weighted voting mechanism that dynamically allocates decision weights based on modality confidence, significantly enhancing classification accuracy. In addition, we also design a multi-loss function module is designed, which combines multiple loss functions for network training and error backpropagation, preserving modality-specific information and avoiding the issue of information annihilation during multimodal fusion. The detailed network structure is shown in Figure \ref{Figure 4}.

\subsubsection{Multiple Loss Module} 
In the multi-loss module of MCFNet, we denote the loss function associated with multimodal feature representation as the primary loss function ${Loss_{H}}$. The loss functions corresponding to the unimodal networks on both sides, which represent text features and image features, are denoted as auxiliary loss functions ${Loss_{T}}$ and ${Loss_{I}}$, respectively. The auxiliary loss functions facilitate the training of the single-modal feature extraction modules on both sides, thereby enhancing the network's ability to process single-modal information. Three fused feature vectors \(O_{T}\), \(O_{I}\), and \(O_{H}\) each undergo independent classification branches followed by the Softmax activation function, resulting in three corresponding loss values. The overall loss function is then computed based on the linear combination of these losses:
\begin{equation}
	Loss = (1 - \gamma )\! \cdot\!Los{s_{H}} + \gamma\! \cdot\!(Los{s_{T}} + Los{s_{I}}),
\label{eq:L15}
\end{equation}

\noindent where $\gamma$ is a hyperparameter that integrates three loss functions, regulating the contribution of the loss functions associated with the single-modal networks on both sides to the overall loss function of the network.

\subsubsection{Multimodal Classification Network} 
We input the feature vectors \(O_{T}\), \(O_{I}\), and \(O_{H}\), obtained through feature extraction and fusion, into the multimodal classification network for fine-grained classification. The network incorporates multiple loss modules, and the multimodal classification module is composed of image classification branch, text classification branch, and interaction classification branch, each accompanied by corresponding Softmax activation functions. For a single sample data $X_m$, the calculation process of the multimodal classification network is computed as follows:
\begin{equation}
	{y_m} = \text{Softmax} ({W_i} \cdot {X_m} + {b_m}),
	\label{eq:L16}
\end{equation}

\noindent where $X_m$ is the multimodal feature representation of data $m$, and $y_m$ is the category probability distribution of data $m$ as predicted by the network. The $W_i$ is the adjustable weight, where \(W_i \geq 0\). The $b_m$ is the bias.

In multi-classification tasks, the cross-entropy loss function is employed for loss calculation and network training. The network's loss function is computed as follows:
\begin{equation}
	Loss(\theta ) = \frac{1}{M}\sum\nolimits_{m = 1}^M {\left( { - \sum\nolimits_{i = 1}^N {{l_m}} \log (y_m^i)} \right)}, 
\label{eq:L17}
\end{equation}

\noindent where $l$ is the real label of data $m$, $M$ represents the batch size of mini-batch, and $N$ is the number of categories for classification.  

To minimize the total loss of the network, we adopt an end-to-end approach for its optimization and learning. The formula for calculating the optimal parameter combination ${\theta ^*}$ is calculated as follows:

\begin{equation}
	{\theta ^*} = \arg \mathop {\min }\limits_\theta  Loss(\theta ).
	\label{eq:L18}
\end{equation}

\section{Experiments}
In this section, we present the specific configuration of the dataset and elaborate on the experimental setup of MCFNet. Additionally, we design a series of comparative experiments and ablation study to validate the effectiveness of MCFNet and the critical role of its core network.  

\label{Section 4}
\subsection{Experimental setup}
\subsubsection{Datasets} 
We use the two public datasets, Con-Text \citep{karaoglu2017text} and Drink Bottle \citep{bai2018integrating}, as the data sources for our network training. The Con-Text dataset comprises 28 categories of street view shops, totaling 24,255 images. Although the text in the images includes salient identifiers like 'Bakery', the presence of decorative semantic content complicates fine-grained image classification. We randomly select 19,404 images for the training set, 2,182 images for the validation set, and 2,910 images for the test set. The Drink Bottle dataset consists of images of 20 different brands and types of alcoholic beverage bottles, totaling 18,488 images. The appearance of different bottles is highly similar, and the brand names vary, which also presents a classification challenge. For the experiment, we divide the dataset into training, validation, and test sets, maintaining the original proportions of 60\%, 10\%, and 30\% for each type, respectively. The details of these two datasets are shown in Table \ref{Table 1}.  

\begin{table}[t]
	\centering
	\setlength{\tabcolsep}{5.5pt}
	\caption{Statistics of the datasets.} 
	\label{Table 1}
	\renewcommand{\arraystretch}{1.2} 
\begin{tabular}{lccc}
	\hline
	\textbf{Dataset} & \textbf{Train} & \textbf{Validation} & \textbf{Test} \\
	\hline
	Con-Text \citep{karaoglu2017text} & 19,404 & 2,182 & 2,910  \\
	Drink Bottle \citep{bai2018integrating} & 11,092 & 1,848 & 5,546  \\

	\hline
\end{tabular}
\end{table}

\subsubsection{Experimental Details} 
This study is conducted on Linux Deepin system and implemented using a framework based on PyTorch 2.1, utilizing NVIDIA GeForce RTX 4080 GPU. During training, the quantization step sizes for both the ImageNet and Drink Bottle datasets are set to 256. The chosen optimizer is AdamW, with a batch size of 8 and a weight decay of $5\times{10^{-4}}$. The training process is conducted over a total of 300 epochs. The learning rate for ALBERT is set to $1\times{10^{-5}}$, for the ViT to $1\times{10^{-4}}$, and for the other modules to $1\times{10^{-4}}$.

Before conducting network training and testing, we adjust all dataset images to 256 × 256 size using a uniform algorithm and normalize the pixel values to a range of [0, 1]. For the text data, we normalize all words to a uniform case and remove the blank spaces.

\subsubsection{Evaluation Index} 
In multimodal classification tasks, accuracy is commonly used as the primary evaluation metric. However, because of class imbalance within the dataset, we also incorporate the harmonic mean of accuracy and recall (F1 score) as a supplementary evaluation metric.

\subsection{Comparison with State-of-the-Art Methods}
To validate the effectiveness of MCFNet, we compare our network with three types of classification models on the Con-Text and Drink Bottle datasets: a) Text unimodal classification models, which include five classic models such as ULMFiT \citep{howard2018universal} and ALBERT \citep{lan2019albert}; b) Image unimodal classification models, which include five classic models such as ResNet50 \citep{he2016deep} and ViT \citep{dosovitskiy2020image}; and c) Image-text multimodal classification models, which include seven models such as MIMN \citep{pi2019practice}, TomBERT \citep{yu2019adapting}, and ViT-BERT \citep{li2021towards}. We also introduce the high-performance multimodal models Flamingo \citep{alayrac2022flamingo}, LLaVa \citep{liu2023visual}, and Qwen-VL \citep{wang2024qwen2}, which have garnered significant attention in recent years, for extended comparisons. The quantitative comparison results of various models are shown in Table \ref{Table 2}.

\begin{table*}[t]
	\centering
	\caption{Classification performance of state-of-the art methods on the Con-Text and Drink-Bottle datasets. } 
	\label{Table 2}
	\renewcommand{\arraystretch}{1} 
	\small 
	\setlength{\tabcolsep}{4pt} 
	\begin{tabular}{llcccccccc}
		\toprule
		\multirow{2}{*}{\textbf{Modal}} & \multirow{2}{*}{\textbf{Model}} & \multicolumn{4}{c}{\textbf{Con-Text}} & \multicolumn{4}{c}{\textbf{Drink Bottle}}\\
		\cmidrule(lr){3-6}\cmidrule(lr){7-10}
		& & \textbf{Acc.(\%)}	&	\textbf{Prec.(\%)}	&	\textbf{Recall(\%)}	&	\textbf{F1(\%)}  &	\textbf{Acc.(\%)}	&	\textbf{Prec.(\%)}	&	\textbf{Recall(\%)}	&	\textbf{F1(\%)} \\
		\midrule
		\multirow{5}{*}{Text}	&	BILSTM \citep{zhang2015bidirectional}	&	86.26	&	85.42	&	85.66	&	83.92  &	84.31	&	82.43	&	81.36	&	81.03	\\
		&	MIARN \cite{tay2018reasoning}	&	86.68	&	85.26	&	86.34	&	84.31	&	84.52	&	82.79	&	82.01	&	81.89	\\
	  	&	ULMFiT \citep{howard2018universal}	&	87.23	&	86.34	&	86.77	&	84.62	&	84.89	&	81.77	&	81.96	&	82.36	\\
		&	ALBERT \citep{lan2019albert}	&	87.51	&	86.47	&	86.56	&	85.01	&	85.27	&	82.54	&	82.88	&	82.75	\\
		&	Flan-T5 \citep{chung2024scaling}	&	88.17	&	87.33	&	87.20	&	85.57 &	85.69	&	83.60	&	83.26	&	83.01	\\
		\midrule
		\multirow{5}{*}{Image}	&	RestNet50 \citep{he2016deep}	&	88.25	&	87.41	&	86.79	&	86.15  &	86.04	&	84.72	&	83.45	&	83.97	\\
		&	API-Net \citep{zhuang2020learning}	&	88.49	&	87.24	&	87.44	&	86.53	&	86.58	&	84.91	&	84.19	&	84.24	\\
		&	ViT \citep{dosovitskiy2020image}	&	89.02	&	88.76	&	88.05	&	87.26	&	86.75	&	83.94	& 84.73	&	84.89	\\
		&	NFNet-F6 \citep{brock2021high}	&	89.73	&	88.57	&	89.11	&	87.64	&	87.13	&	85.28	&	85.22	&	85.63	\\
		&	TransFG \citep{he2022transfg}	&	90.42	&	89.69	&	89.76	&	87.96	&	87.64	&	85.76	&	84.69	&	85.74	\\
		\midrule
		\multirow{8}{*}{Text-Image}	&	Concat \citep{schifanella2016detecting}	&	90.66	&	89.45	&	89.63	&	88.03	&	87.99	&	85.02	&	85.81	&	85.96	\\
		&	MIMN \citep{pi2019practice}	&	90.97	&	89.23	&	88.34	&	88.39	&	88.21	&	86.75	&	86.33	&	86.45  \\
		&	TomBERT \citep{yu2019adapting}	&	91.35	&	90.48	&	89.21	&	89.44	&	88.47	&	86.41	&	86.58	&	86.88	\\
		&	ViT-BERT \citep{li2021towards}	&	91.78	&	90.60	&	88.26	&	88.57	&	89.25	&	87.03	&	87.13	&	87.64	\\
		&	Flamingo \citep{alayrac2022flamingo}	&	92.56	&	91.57	&	89.73	&	90.03  &	89.63	&	87.45	&	87.64	&	88.11	\\
		&	LLaVa \citep{liu2023visual}	&	92.67	&	90.98	&	90.04	&	89.26   &	90.33	&	88.06	&	88.69	&	88.67	\\
		&	Qwen2-VL \citep{wang2024qwen2}	&	92.89	&	91.07	&	90.36	&	89.78	&	91.89	&	88.34	&	89.20	&	89.45	\\
		&	\textbf{MCFNet}	&	\textbf{93.14}	&	\textbf{92.52}	&	\textbf{91.22}	&	\textbf{90.37}	 &	\textbf{92.23}	&	\textbf{90.56}	&	\textbf{89.97}	&	\textbf{90.02}	\\
		\bottomrule
	\end{tabular}
\end{table*}

By comparing the experimental data of all modal classification models presented in Table \ref{Table 2}, our proposed MCFNet demonstrates superior performance on both benchmark datasets. Because of the complementary nature of information from different modalities, multimodal classification models generally achieve higher classification accuracy than single-modal classification models through efficient cross-modal information fusion. Specifically, on the Con-Text dataset, MCFNet's classification accuracy is 4.97\% higher than that of the best single-modal text model, ALBERT, and the F1 metric improves by 4.8\%. Compared to the best single-modal image model, ViT, MCFNet's accuracy increases by 2.83\%, and the F1 metric improves by 2.41\%. This enhancement in accuracy validates the effectiveness of multimodal collaborative modeling. MCFNet achieves deep semantic alignment of text and image features through hybrid attention mechanisms, thereby enabling the model to exhibit stronger discriminative feature learning capabilities in fine-grained classification tasks.

To comprehensively evaluate the network's performance, we conducted a comparative analysis with other mainstream multimodal fine-grained classification models on two benchmark datasets: Con-Text and Drink Bottle. The experimental results demonstrate that the MCFNet achieved optimal performance across all evaluation metrics. Compared to the second-ranked model, MCFNet improves classification accuracy by 0.25\%, precision by 1.45\%, and F1 score by 0.58\% on the Con-Text dataset. On the Drink Bottle dataset, it increased classification accuracy by 0.34\%, precision by 2.22\%, and F1 score by 0.57\%. These results fully substantiate the advanced capabilities of MCFNet in multimodal fine-grained image classification tasks.

Furthermore, we conduct a specialized comparison of MCFNet with TomBERT \citep{yu2019adapting} and ViT-BERT \citep{li2021towards}, both of which are based on pre-trained architectures. On the Con-Text dataset, MCFNet achieves an F1 score increase of 1.36\% over ViT-BERT and 1.79\% over TomBERT. On the Drink Bottle dataset, MCFNet achieves an F1 score increase of 2.98\% over ViT-BERT and 3.76\% over TomBERT. This comparative analysis demonstrates that MCFNet, by introducing a multimodal collaborative regularization method, effectively mitigates the prevalent overfitting issue in traditional pre-trained model transfer learning, significantly enhancing the model's generalization capability while maintaining training efficiency.

\begin{figure*}[t]
        \raggedright
	\centering
	 \subfigure[Con-Text]{
        \includegraphics[scale=0.35]{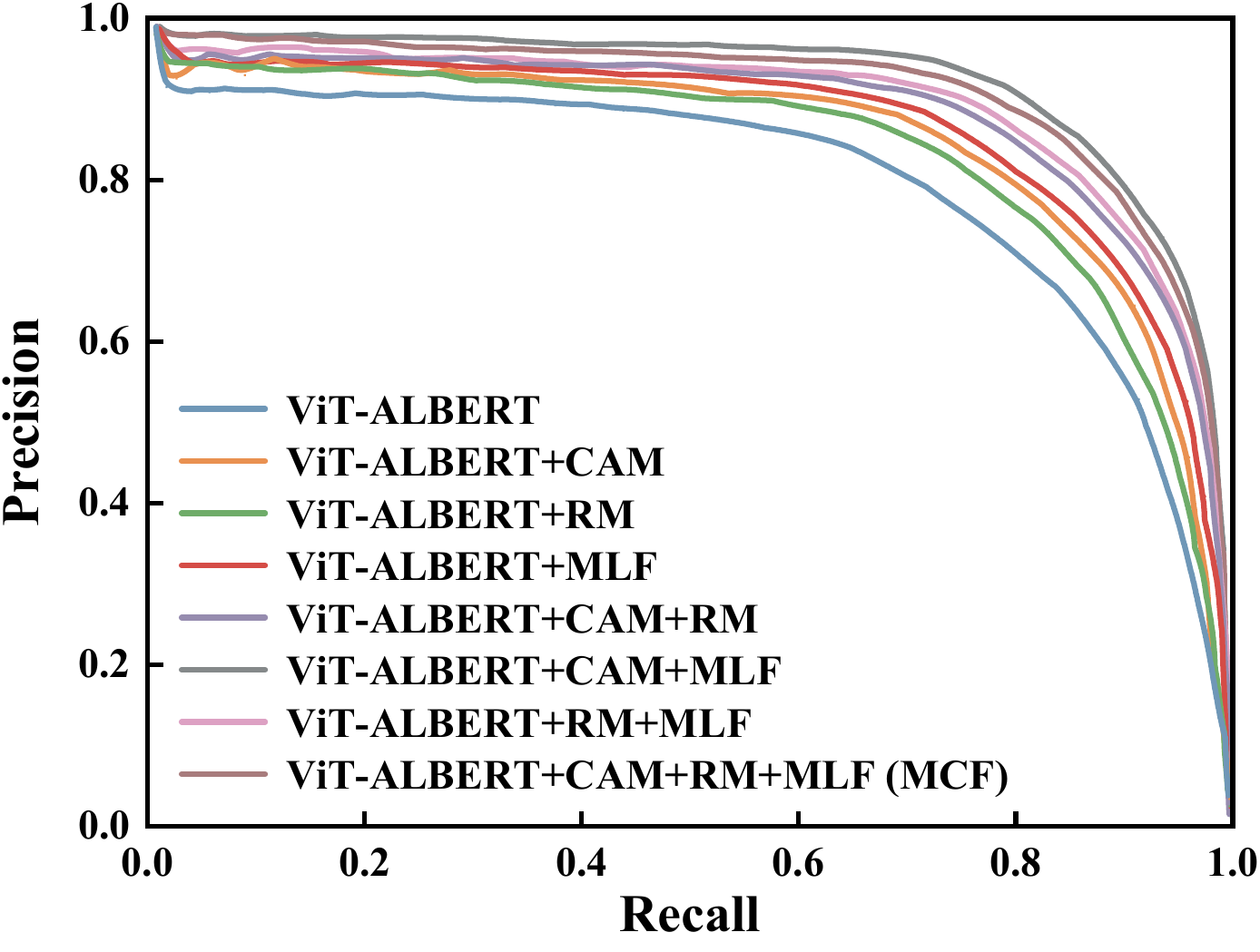}
        \label{Figure 5-(a)}
    }
     \hspace{0.2cm}  
    \subfigure[Drink Bottle]{
        \includegraphics[scale=0.35]{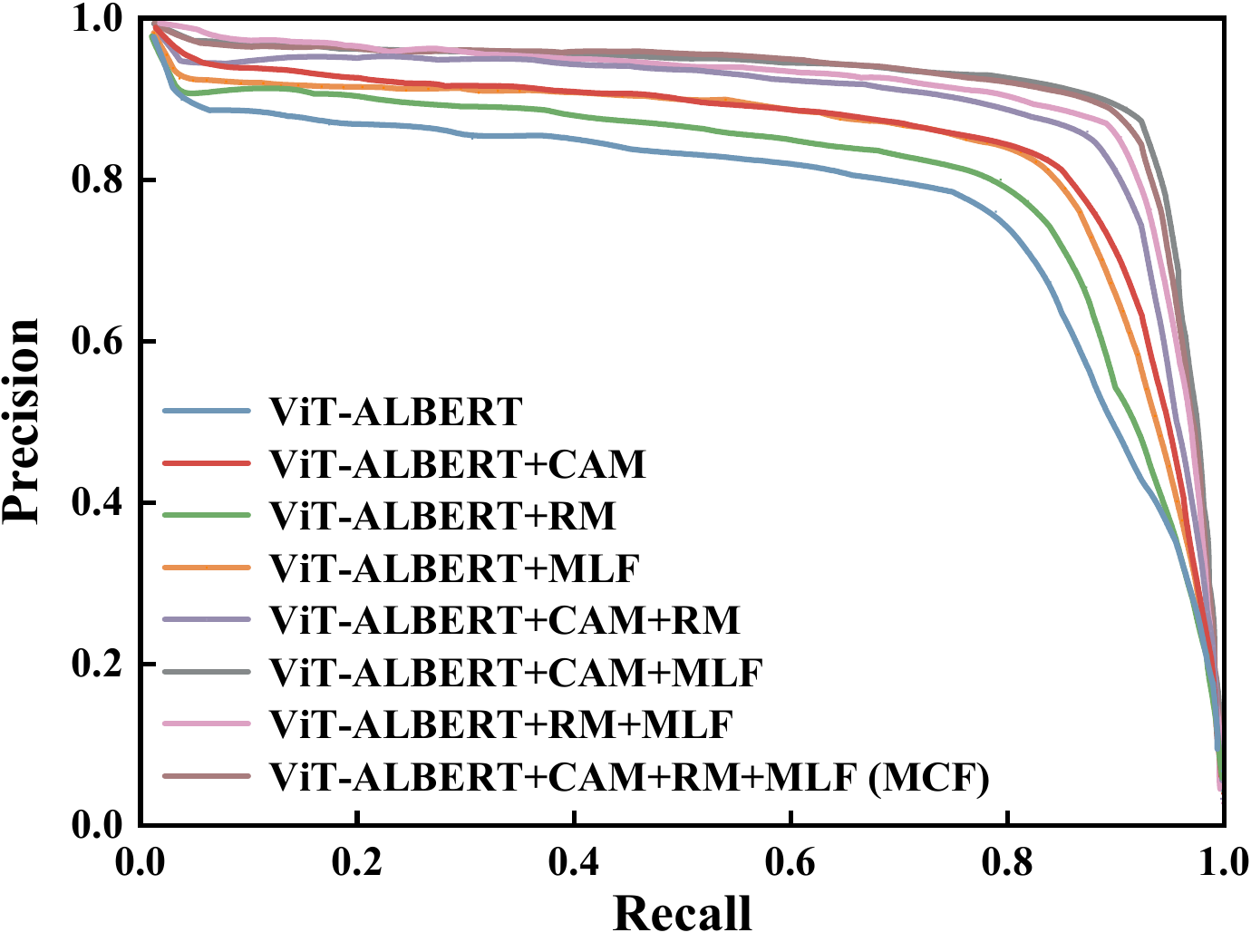}
        \label{Figure 5-(b)}
    }
    \caption{PR curves of different ablation methods.}
    \label{Figure 5}
    \vspace{-0.5cm}
\end{figure*}

\subsection{Ablation Study}
\subsubsection{Component Analysis}
In this section, we perform ablation experiments on the hybrid attention mechanism network (HAM), regularization method (RM), and multiple loss functions module (MLF) incorporated within the MCFNet using the Con-Text and Drink Bottle datasets. The experiments investigate the effectiveness of these three modules and analyze the hyperparameters involved. Relative optimal results are shown in bold.

We select ViT-ALBERT as the baseline model and incorporate HAM, RM, and MLF in various cross-combinations for performance comparison. Table \ref{Table 3} shows the results of our indicators. The results indicate that on the Con-Text dataset, the introduction of HAM, RM, and MLF resulted in increases in classification accuracy of 0.59\%, 0.16\%, and 0.45\%, respectively, with corresponding improvements in F1 metric of 0.63\%, 0.22\%, and 0.39\%. On the Drink Bottle dataset, classification accuracy increased by 0.81\%, 0.32\%, and 0.49\%, respectively, with the F1 metric with corresponding improvements of 0.77\%, 0.48\%, and 0.59\%. Furthermore, on the Con-Text dataset, the combination of HAM and MLF achieved a classification accuracy of 92.91\%, representing an improvement of 0.52\% compared to using HAM alone and a 0.66\% increase over using MLF alone. This synergistic effect was also significant on the Drink Bottle dataset, where the combined model achieved accuracy improvements of 1.20\% and 1.52\% compared to the individual modules, respectively. These results demonstrate the effectiveness of incorporating the aforementioned three modules in MCFNet for improving fine-grained classification tasks.

\begin{table}[t]
	\centering
	\caption{Experimental results of ablation of components in MCFNet.} 
	\label{Table 3}
	\renewcommand{\arraystretch}{1.3} 
	\small 
	\setlength{\tabcolsep}{2.5pt} 
	\begin{tabular}{ccccccc}
		\toprule
		\multicolumn{3}{c}{\textbf{Module}} & \multicolumn{2}{c}{\textbf{Con-Text}} & \multicolumn{2}{c}{\textbf{Drink Bottle}} \\
		\cmidrule(lr){1-3} \cmidrule(lr){4-5} \cmidrule(lr){6-7}
		\textbf{HAM}  & \textbf{RM} & \textbf{MLF}	& \textbf{Accuracy(\%)}	&	\textbf{F1(\%)}	& \textbf{Accuracy(\%)}	& \textbf{F1(\%)} \\
		\midrule
		-	&	-	&	-	&	91.80	&	88.62	&	89.94	&	88.27	\\
		\Checkmark	&	-	&	-	&	92.39	&	89.25	&	90.75	&	89.04	\\
		-	&	\Checkmark	&	-	&	91.96	&	88.84	&	90.26	&	88.75	\\
		-	&	-	&	\Checkmark	&	92.25	&	89.01	&	90.43	&	88.86	\\
		\Checkmark	&	\Checkmark	&	-	&	92.83	&	89.76	&	91.64	&	89.67	\\
		\Checkmark	&	-	&	\Checkmark	&	92.91	&	89.83	&	91.95	&	89.81	\\
		-	&	\Checkmark	&	\Checkmark	&	92.77	&	89.68	&	91.47	&	89.23	\\
		\Checkmark	&	\Checkmark	&	\Checkmark	&	\textbf{93.14}	&	\textbf{90.37}	&	\textbf{92.23}	&	\textbf{90.02}	\\
		\bottomrule
	\end{tabular}
	\begin{tablenotes}[flushleft]
	\item \checkmark means the module is added, $\textbf{-}$ means the module is removed.
	\end{tablenotes}
\end{table}

The precision-recall (PR) curves of the ablation experiments are shown in Figure \ref{Figure 5}. On both benchmark datasets, the trends of the PR curves for different module combinations are highly consistent with the quantitative classification accuracy results presented in Table \ref{Table 3}. The results demonstrate that when any single module (HAM/RM/MLF) is ablated from the complete MCFNet, the network's classification accuracy significantly decreases on both datasets. The absence of HAM leads to the largest accuracy drop on the Con-Text dataset (up to 1.2\%), while MLF exhibits the strongest necessity on the Drink Bottle dataset (with an accuracy drop of 0.8\%). This phenomenon fully validates the collaborative effectiveness of each module in enhancing multimodal representation capabilities, demonstrating that all components in the MCFNet proposed in this study are essential configurations for achieving optimal performance.

\begin{figure*}[t]
        \raggedright
	\centering
	 \subfigure[Comparison of accuracy with different $\gamma$ values on the Con-Text and Drink Bottle datasets.]{
        \includegraphics[scale=0.35]{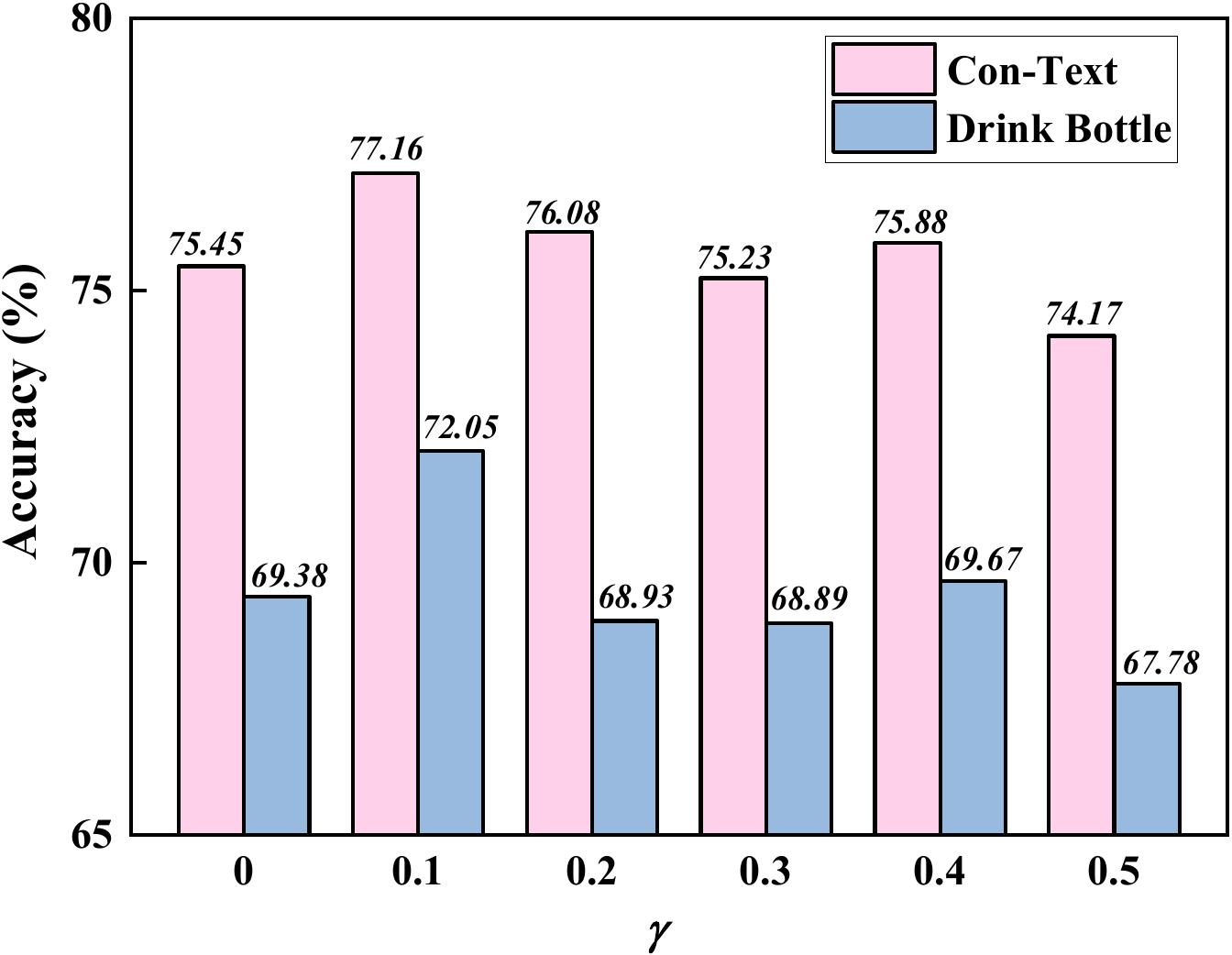}
        \label{Figure 6-(a)}
    }
     \hspace{0.2cm}  
    \subfigure[Comparison of F1 with different $\gamma$ values on the Con-Text and Drink Bottle datasets.]{
        \includegraphics[scale=0.35]{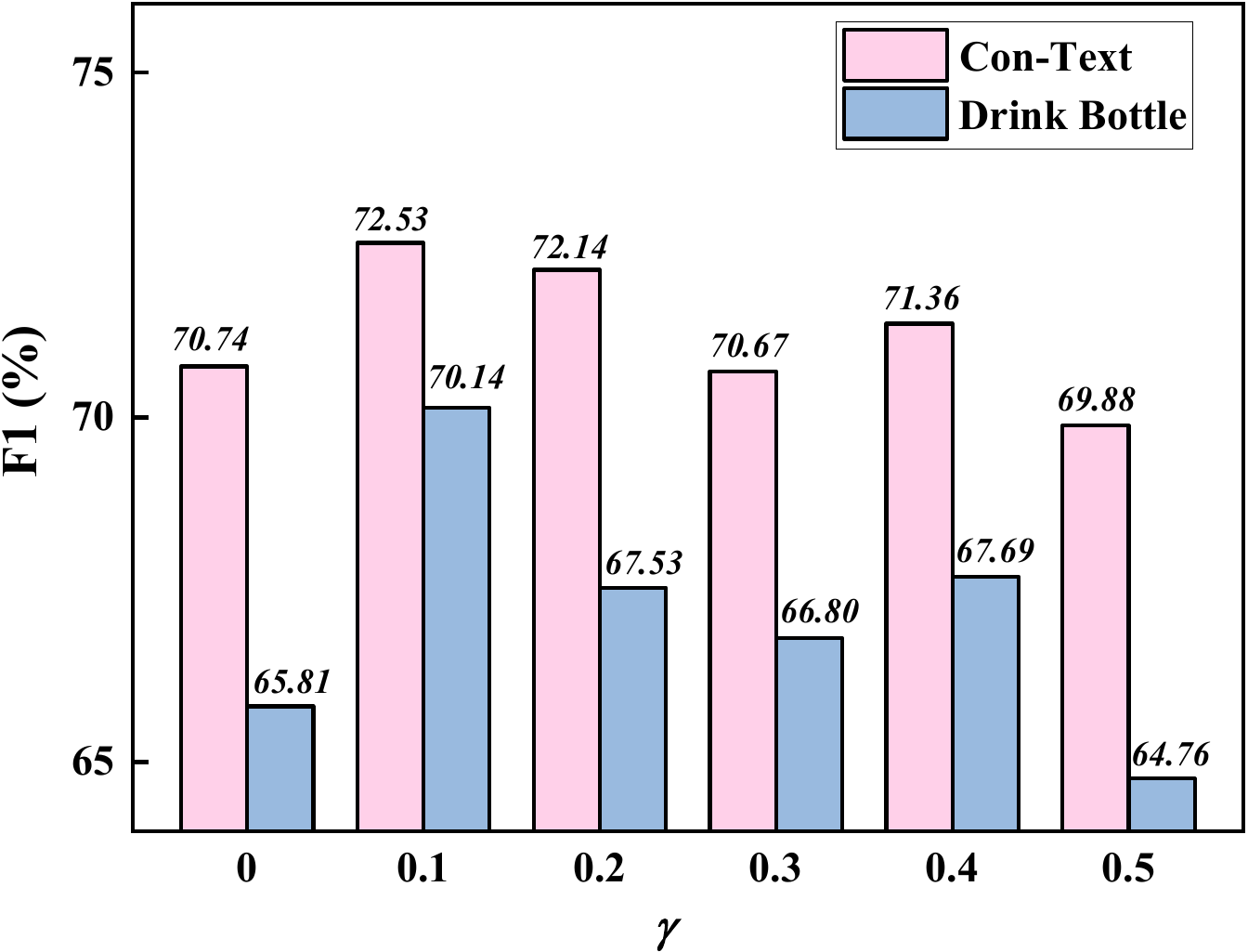}
        \label{Figure 6-(b)}
    }
    \caption{Experimental results corresponding to different values of the trade off hyperparameter $\gamma$ .}
    \label{Figure 6}
    \vspace{-0.5cm}
\end{figure*}

\subsubsection{Analysis of the Loss Function Hyperparameter}
According to the description of MLF in Section \ref{section 3.4} and Eq. (\ref{eq:L15}), $\gamma$ regulates the influence of the loss functions from the two single-modal networks on the total loss function of MCFNet. A value of $\gamma=0$ indicates that MCFNet eliminates the loss function associated with the single-modal feature representation, retaining only the loss function related to the multi-modal feature representation. As $\gamma$ decreases, the influence of the loss functions from the two single-modal networks on the total loss function diminishes, and vice versa. Since MCFNet prioritizes the multiple loss functions, values of 0, 0.1, 0.2, 0.3, 0.4, and 0.5 are selected as candidate parameters for $\gamma$ in the experiments, while all other experimental parameters remain constant. The specific results of the ablation experiments are presented in Figure \ref{Figure 6}, with the relatively optimal results highlighted in bold.  
The results show that when MLF is employed ($\gamma>0$), MCFNet achieves optimal performance across all metrics on both datasets at $\gamma=0.1$. Conversely, when $\gamma=0$, which signifies the removal of the MLF, the performance metrics of MCFNet on both datasets are inferior to those observed at $\gamma=0.1$. This finding underscores the importance and effectiveness of MLF.

\subsubsection{Analysis of the Hybrid Attention Network}
To demonstrate the advantages of our proposed hybrid attention network, we compare it with two other mainstream multimodal attention mechanism modules. The hybrid attention network in MCFNet is a computationally efficient operation that can accurately capture the intrinsic relationships and complementary information among different modal features, thereby significantly enhancing the effectiveness of multimodal data fusion. Under the configuration of our proposed MCFNet, we conducte extensive comparisons with \textit{Merged Attention} and \textit{Interaction Encoder}, with the experimental results presented in Table \ref{Table 5}. The results indicate that although our proposed network has slightly more parameters than the previous two methods, it only requires 11.25 ms of processing time, showcasing higher computational efficiency. Furthermore, it achieves an excellent accuracy of 93.14\%, which is significantly higher than the other two methods. This demonstrates that our network can achieve more precise fine-grained image classification.

\begin{table}[t]
    \centering
    \caption{Experimental results corresponding to different multimodal attention modules of MCFNet on Con-Text.}
    \label{Table 5}
        \renewcommand{\arraystretch}{1.2} 
    \setlength{\tabcolsep}{0.7pt} 
       \begin{tabular}{lccc}
\toprule
\multirow{2}{*}{\textbf{Method}} & \multicolumn{3}{c}{\textbf{Con-Text}} \\ 
\cmidrule{2-4}
 & \textbf{Parameter (M)}$\uparrow$ & \textbf{Time (ms)}$\downarrow$ & \textbf{Accuracy (\%)}$\downarrow$  \\ 
\midrule
\textit{Merged Attention} & \textbf{14.67
}& 19.36 & 89.67 \\ 
\textit{Interaction Encoder} & 19.89 & 14.38 & 90.31 \\ 
\textbf{ours} & 22.34 & \textbf{11.25} & \textbf{93.14} \\ 
\bottomrule
\end{tabular}
\end{table}

\subsubsection{Analysis of the Regularization Methods}
To demonstrate the effectiveness of the proposed regularization method, we conducted ablation experiments on the three regularization methods discussed in Section \ref{section 3.3}. The experimental results are shown in Table \ref{Table 6}. After removing dropout regularization, the network exhibite a certain degree of performance decline across various metrics on the Con-Text and Drink Bottle datasets. This indicates that dropout regularization helps prevent network overfitting and enhances the network's generalization capability. Similarly, the removal of the elastic network resulted in a decrease in network performance. It is evident that the L1 and L2 regularization method within the elastic network can impose constraints and optimize the network's feature representations, thereby improving the network's adaptability to different tasks. The removal of the hybrid attention mechanism led to the poorest performance of the network across both tasks. On the Con-Text dataset, the accuracy dropped to 88.19\%, precision to 87.52\%, and F1 score to 86.46\%; on the Drink Bottle dataset, accuracy fell to 87.66\%, precision to 85.51\%, and F1 score to 85.20\%. This demonstrates that the hybrid attention mechanism effectively captures the intermodal feature correlations, playing a crucial role in enhancing the network's performance. In summary, the three proposed regularization methods are indispensable for improving the network's performance on the relevant tasks.

\begin{table}[t]
      \centering
    \caption{The ablation experimental results of different regularization methods.}
    \label{Table 6}
        \renewcommand{\arraystretch}{1.2} 
    \setlength{\tabcolsep}{0.8pt} 
    \begin{tabular}{lcccccc}
        \toprule
        \multirow{2}{*}{\textbf{Method}} & \multicolumn{3}{c}{\textbf{Con-Text}} & \multicolumn{3}{c}{\textbf{Drink Bottle}} \\
        \cmidrule(lr){2-4} \cmidrule(l){5-7}
        & \textbf{Acc.(\%)} & \textbf{Prec.(\%)} & \textbf{F1(\%)} & \textbf{Acc.(\%)} & \textbf{Prec.(\%)} & \textbf{F1(\%)} \\
        \midrule
        MCFNet & 93.14 & 92.52 & 90.37 & 92.23 & 90.56 & 90.02\\
        w/o Dropout & 92.25 & 91.02 & 88.97 & 91.74 & 89.62 & 88.68 \\
        w/o Elastic Net. & 91.03 & 88.76 & 87.96 & 91.56 & 90.28 & 89.73 \\
        w/o Hybrid Atte. & 88.19 & 87.52 & 86.46 & 87.66 & 85.51 & 85.20 \\

        \bottomrule
    \end{tabular}
\end{table}

\begin{figure*}[t]
	\centering
	\includegraphics[scale=0.74
	]{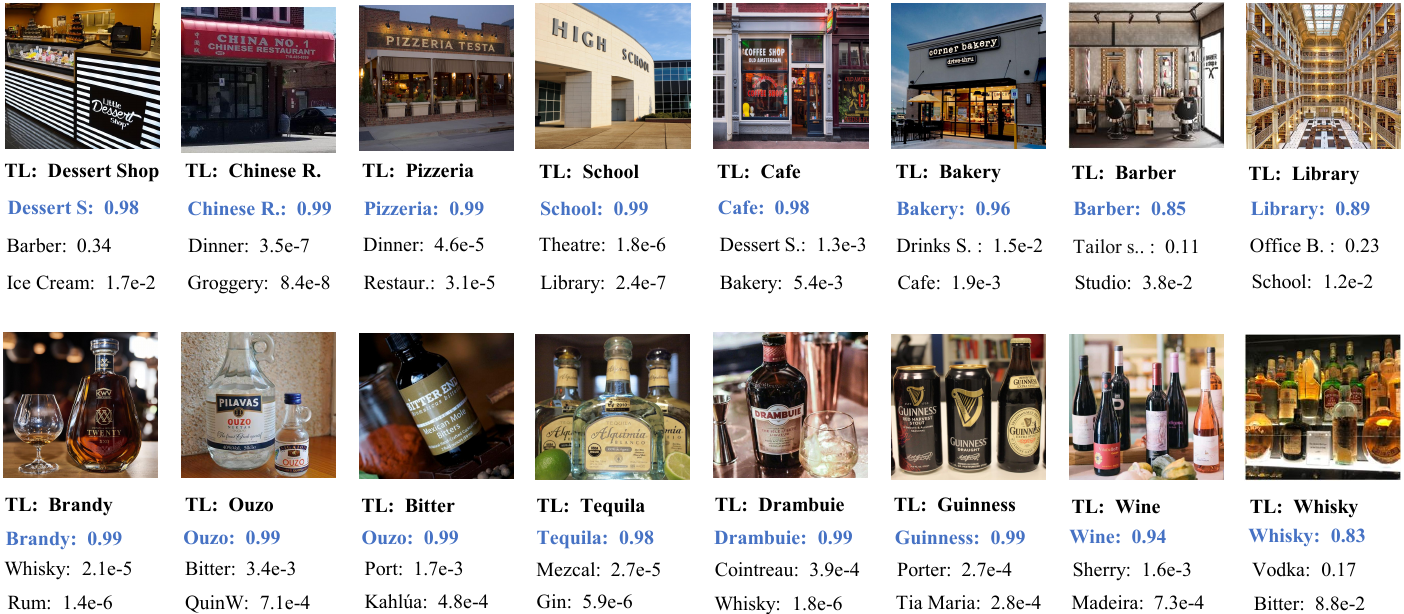}
	\caption{Classification prediction. The figure illustrates the true label of a category, along with the top three probability-ranked results from the network's classification of this category. In this representation, blue is used to indicate correct predictions. }
	\label{Figure 7}
    \vspace{-0.5cm}
\end{figure*}

\subsection{Qualitative Results}
The qualitative results of the fine-grained image classification task are shown in Figure \ref{Figure 7}. The results demonstrate that our network effectively integrates visual and textual modal information from a single image to achieve accurate classification. For example, in the first row of the Con-Text dataset, the network accurately identifies the scenes of "Barber," "Bakery," and "Desert Shop," capturing both the street scene layout and the visual features of shop signs, while also incorporating specific brand information from the textual descriptions. Similarly, for the Drink Bottle dataset, MCFNet emphasizes visual cues such as bottle contours and label patterns, working in conjunction with brand information from the text annotations to achieve precise classification at a fine-grained level. This fully demonstrates MCFNet's generalization and robustness across different scenarios.

\section{Conclusion}
\label{Section 5}
In this paper, we propose a multimodal collaborative fusion framework named MCFNet for fine-grained semantic classification tasks. To address the overfitting problem associated with the transfer of pre-trained models, we introduce regularization methods tailored to different modalities. To enhance the semantic relationship between text and image features, we design a multimodal regularization integrated fusion module that achieves bidirectional alignment of visual and textual features. Additionally, we devise a multimodal decision classification module that effectively reduces prediction bias caused by missing unimodal information through dynamic adjustment of decision weights. Experiments conducted on two datasets validate the robustness of MCFNet, demonstrating significant improvements in classification accuracy compared to existing methods. In the future, we expect to explore the application of lightweight networks to expand the use of MCFNet in more practical scenarios.

\printcredits
\section*{Declaration of competing interest}
The authors declare that they have no known competing financial interests or personal relationships that could have appeared to influence the work reported in this paper. 

\section*{Data availability}
Data will be made available on request.

\section*{Acknowledgments}
This work was supported in part by the Key Research Project of Jiangsu Province, China under Grant BE2023019-3.
\balance
\bibliographystyle{model1-num-names}
\bibliography{my-refs}



\end{document}